# Fast Robot Arm Inverse Kinematics and Path Planning Under Complex Static and Dynamic Obstacle Constraints

David W. Arathorn
Dept of Electrical and Computer Engineering, Montana State University.
dwa@giclab.com, dwa@cns.montana.edu

**Abstract**

Described here is a simple, reliable, and quite general method for rapid computation of robot arm inverse kinematic solutions and motion path plans in the presence of complex obstructions. The method derived from the MSC (map-seeking circuit) algorithm, optimized to exploit the characteristics of practical arm configurations.   The representation naturally incorporates both arm and obstacle geometries.  The consequent performance on modern hardware is suitable for applications requiring real-time response, including smooth continuous avoidance of dynamic obstacles which impinge on the planned path during the traversal of the arm.  On high-end GPGPU hardware computation of both final pose for an 8 DOF arm and a smooth obstacle-avoiding motion path to that pose takes approximately 200-300msec depending on the number of waypoints implemented. The mathematics of the method is accessible to high school seniors, making it suitable for broad instruction. *[Note: This revision includes a general compute strategy for paths from arbitrary pose to arbitrary pose and a compute strategy for continuous motion midcourse avoidance of dynamic obstacles.]*

**Introduction**

The term "inverse kinematics" as applied to robotic arm manipulators denotes the problem of determining the angles of each of the joints which, given a root position in 3-space, will locate the distal end of the last arm segment (or end effector) at a specified location in 3-space.  There is a widely used method for solving inverse kinematic (IK) problems of this sort, which, for the audience of this paper, need not be described here.  A limitation of this method is that when the IK problem needs to be solved for an arm operating in the presence of complex obstacles, the process of solution becomes awkward and computationally very expensive.  A common approach to these difficulties involves precomputing a discretized space of arm configurations which avoid the fixed obstacles and self-collisions,the discretized space then organized as a graph of nearby configurations. During operation the algorithm involves finding a path through this graph from a node near the starting configuration to the node nearest the ending configuration.  The limitation here is that the practical number of nodes limits the resolution of the configuration space so that constricted motion paths that require small joint angle adjustments may not be discovered.  Of course, any obstacles not accounted for in the precomputed restriction of the configuration space must be tested on the fly.  Methods in the above category will not be discussed here because the method described here derives from completely different origins.  The underlying principle is that for clean solution of an inverse problem under constraints the representation has to naturally incorporate those constraints, no matter how complex they may be in practice.

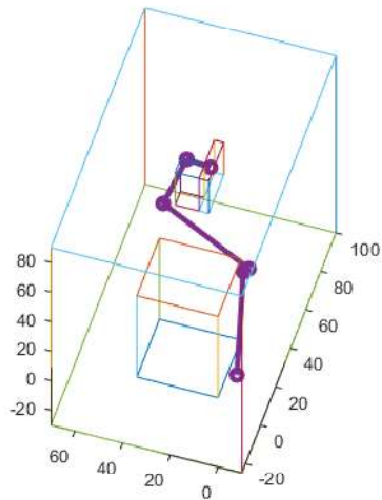

**Fig 1: 8DOF Arm "Reach Pose" Solution Amid Obstacles (shown as blocks)**

The practical use of robotic arms requires not only the solution of the articulation which will bring its working end to the target (Fig 1): here referred to as the "reach pose" or final pose to distinguish it from the sequence of articulations which take it there from its initial pose. It also requires a path of motion that takes it from its starting articulation to the reach pose. When operating in the presence of obstacles, determining this motion path requires solving for a smooth sequence of articulations also constrained by the same obstacles.

This technical paper describes a method which, relying on a technique originating in another computational domain, gracefully and quickly solves the full problem described above: IK for both the reach pose and the motion path in the presence of arbitrarily complex obstacles. The method, when deployed on modern parallel computational hardware, can solve the full reach pose and path planning problem for an 8 degree-of-freedom arm, in a couple of hundred milliseconds for most normal obstacle situations. This speed allows accommodation of the arm to dynamic as well as static obstacles.

The method invoked for this purpose derives from the MSC (map-seeking circuit) algorithm [1]. MSC is a method for solving inverse problems which can be posed as a composition of transformations. The conventional MSC variant has been used extensively for machine vision, in which recognizing or interpreting the image of an object involves determining the sequence of visual transformations which map the image to a 3-D representation or model of the object. The conventional form of MSC can also be applied to IK problems. For robotic arms, the sequence of transformations (in effect, translations) to be determined are the sequence of mappings that take the root to the distal location of the first arm segment, the latter location to the distal location of the next arm segment, and so forth until the last transformation takes the distal location of the last segment (or end effector) to the location of the target. The variant version of MSC described here solves the same problem, using a very similar representation, but exploits the specific characteristics of the arm IK problem for computational efficiency.

The means by which conventional MSC obtains its computational efficiency is by formation of superpositions of transforms and the "collapsing" of these superpositions by iterative convergence. The variant described here, while using essentially the same representation of the problem, bypasses the method of solution of the conventional form for a much more readily understood highly pruned search. For those interested in the more general, conventional form of MSC, references are provided in the Appendices. (These are not necessary for understanding the present variant.)

The IK problem for a conventional 8DOF arm stated as a composition of transformations is

$$\mathbf{p}^{end} = L^4 \cdot t_l^{seg\,4} \circ L^3 \cdot t_k^{seg\,3} \circ L^2 \cdot t_j^{seg\,2} \circ L^1 \cdot t_i^{seg\,1} \left( \mathbf{p}^{root} \right) \qquad t_j^{seg} \in T \qquad \text{(eq 1)}$$

where the knowns are $p^{end}$, the end effector locus, and $p^{root}$, the arm root locus. The unknowns are the sequence of transformations, $t$, that take $p^{root}$ to $p^{end}$. In conventional MSC all the transformations are translations represented as vectors of the length $L^{seg}$ of the corresponding arm segment whose direction corresponds to 2 degrees of freedom for the arm segment direction *in the global frame*. The discretized set of vectors for each segment, indexed by $i, j, k, l$ belong to a set of unit direction vectors, T, called the "quiver." The variant method uses essentially the same representation. Since the transformations here are translations they are implemented by vector additions. So eq. 1 is equivalent to

$$\mathbf{p}^{end} = \left( L^4 \cdot \mathbf{t}_l^{seg\,4} \right) + \left( \left( L^3 \cdot \mathbf{t}_k^{seg\,3} \right) + \left( \left( L^2 \cdot \mathbf{t}_j^{seg\,2} \right) + \left( \left( L^1 \cdot \mathbf{t}_i^{seg\,1} \right) + \mathbf{p}^{root} \right) \right) \right)$$

However, the compositional notation of eq.1 is easier to read, so will be used henceforth.

In both conventional MSC and the variant method the pose solution is a vector representation of the arm segment positions and orientations in global space. This facilitates efficient testing on the fly for collision with obstacles, whose representation is naturally in the global space. The vector representation is readily converted to a joint angle representation to control the actuators of most physical arms. Though the vector representation co-locates two degrees of freedom in the same joint, in physical arms the axial rotation does not have to be co-located with flex rotation. Offset joints require slightly different calculation, as discussed later.

**The representations of the arm and obstacles**

MSC is a selection process. It selects a sequence of discreet transformations from sets of families of transformations, such that the composition of the selected sequence of mappings transforms one input parameter to another input parameter. The parameters may be thought of as variables or patterns. In conventional MSC the parameters are vectors representing space, and the mappings move the elements of the vectors in prescribed ways corresponding to geometric transformations. So, in a conventional MSC, the root location of the arm and its target would both be represented as a single

non-zero element in a 3 dimensional array, and the MSC would determine a sequence of mappings which move the non-zero element in stages through 3-space to the target location. As mentioned above, this is a discretized representation of the IK problem: both space and the mappings must be discretized, so the initial solution is approximate. In this representation it is easy to see how obstacles can be represented in the same form. In a 3D array of the same dimensions as those which represent the arm segment endpoint locations, all the space occupied by obstacles is designated by a particular value, say zero, and all the free space is designated by another value, say 1. This 3D obstacle array can then be used as a mask to block any solutions which locate an arm segment in the voxels occupied by the obstacles.

The voxel representation of obstacles allows capture of obstacles in real-time by cameras that produce depth images. For reasons that will become apparent, obstacle volumes need not be filled. Dilation of the "skin" of obstacles, as captured by depth cameras, to a thickness greater than the sampling or waypoint interval, to be discussed later, suffices. However, obstacle capture is outside the scope of this discussion.

In conventional MSC as applied to IK, the mappings which represent arm segment positions are computed numerically from the coordinates of the spatial representation and then rounded back into the spatial representation just described for each step of the solution. (This derives from the machine vision application in which the array representation must be preserved for images.) In the variant to be described here, arm locations remain represented numerically, while obstacles are represented spatially, as voxels, as just described, and the method maps between the two representations on the fly to achieve the same masking as described above. Despite the numerical representation, the process *mostly* remains a selection among discrete mappings. The word "*mostly*" is significant. The difference, as will be seen, allows the derived method to compute exact IK solutions despite discretizations of the mappings. For reasons related to "mostly" this derived method is termed "gap map-seeking" or gMS. It is not termed gMSC because the algorithm does not have the "circuit" characteristics of standard MSC.

**Discretization of the segment articulations**

It has been found from experience that the best discretization for MSC of any 2DOF rotation is a so-called "tiling of the sphere." The objective is to obtain a set of unit vectors from the origin which touch a sphere such that the sampling density of the sphere surface is approximately constant.

One way to accomplish this is to subdivide the equator into *n* divisions of azimuth, and use the great circle distance (or an approximation) between divisions along the equator as the approximate spacing for most of the other latitudes. This results in decreasing numbers of generated vectors as the poles are approached but a minimum number is maintained at high elevations. The elevation angles, or latitudes, are equidistant, as usual. The angle information is discarded, but the addressing of the vector set is by two indices: [elevation order, azimuth order], with a different number of azimuth indices for each elevation. The result is a *quiver* of unit vectors from the sphere center to the points on the sphere surface of approximately uniform over most of the sphere.

A single quiver is used to establish the joint rotation choices for the first and second and fourth (if necessary) segments' articulation mappings, and the mappings themselves are simply the quiver unit vectors $\mathbf{r}_{i,j}$ multiplied by the segment length $l_s$ and restricted to some subset by other constraints. In practice elevation and equatorial azimuth angles of 1 or 2 degrees have proven effective, though smaller angle increments can be used at the cost of compute time. The quiver also has the side benefit of limiting the number of nearly equivalent solutions.

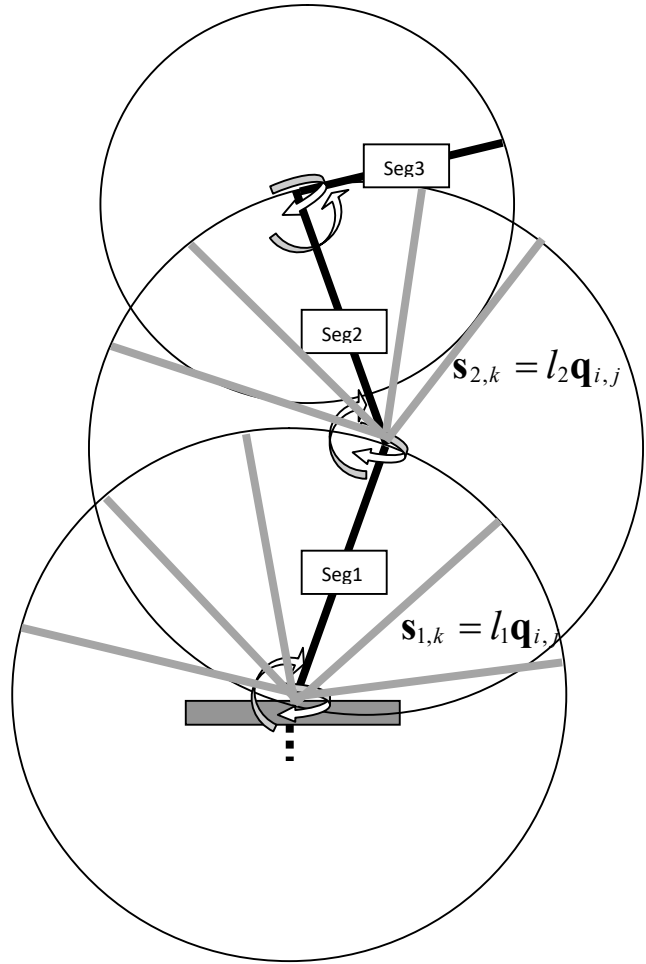

**Fig 2: Segments 1 and 2 quiver generated mappings.** The endpoints for segment 1 and segment 2 are $\mathbf{p}_{1,k1} = \mathbf{s}_{1,k1} + \mathbf{p}_0$ and $\mathbf{p}_{2,k2} = \mathbf{s}_{2,k2} + \mathbf{p}_1$ respectively. ($\mathbf{p}_0$ is root position)

Note in Fig 2 that the third segment is not shown as part of a quiver generated mapping set. This will be explained in the next section.

**The gap in gMS**

The selection process for an 8DOF arm will be described. This adds a flexion joint (corresponding approximately to the knuckle) to the standard 7 DOF arms commercially available, and allows IK solution all the way to the angle of attack of a "finger" in a humanoid hand. The path planning step will use 6DOF calculations for efficiency. High DOF arms are necessary for obstacle avoidance and dexterity in any reasonably complex task scenario.

As in conventional MSCs, there are forward and backward pathways in the gMS. However, unlike the conventional form they do not go from end to end. The forward path is defined to originate at the root and the backward path to originate at the target locus. The forward path for 8DOF has only two layers (in old MSC terminology) or stages: those for the first and second segments. The mappings for these are defined by the quiver, described above. In practice only part of the quiver may be available due to joint angle restrictions (e.g the first segment from the root may for some arm architectures be restricted to the upper hemisphere of the quiver, or less). The backward path originates at the end effector target. It represents the set of permissible angles of attack of the $4^{th}$ segment (end effector or fingers) to the target object. If the application requires a single angle of attack, as in the case of a drill bit, then there is a single vector representing the backward path, but in manipulation tasks there may be a cone of permissible angles. In the latter case, a cone of vectors out of the quiver will constitute the set of available mappings for the $4^{th}$ segment. The backward mappings therefore form either a point, or points on a section of sphere bounded by the permissible cone.

To simplify the discussion the axial rotation of this final segment is ignored here. For a four segment arm eight degrees of freedom are assumed. To allow full axial rotation of the fourth segment an additional joint is required because obstacle constraints will often preclude redundancy of the arm to provide more than limited axial rotation of the fourth (end effector or tool) segment.

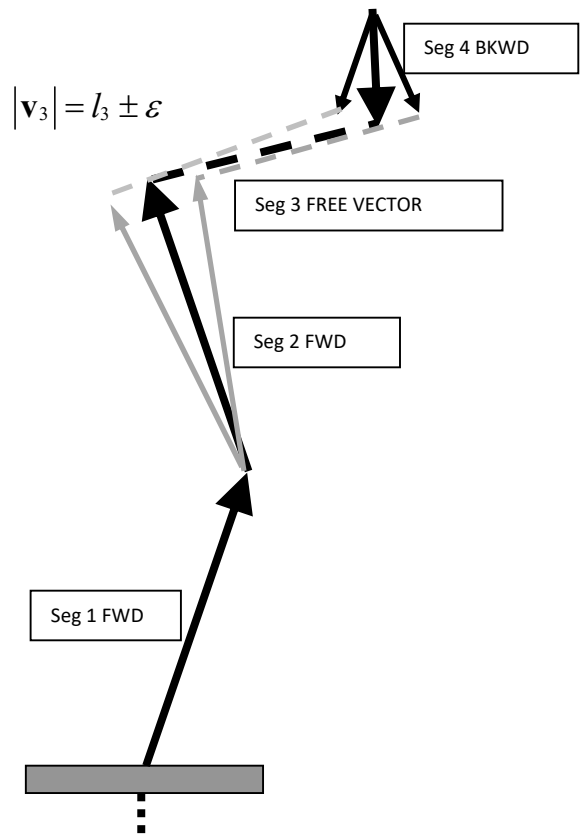

**Fig. 3: gMS forward and backward (cone) mappings, and free vector in gap.**

Notice that there is no reference to a mapping set for segment 3. This is the *gap* in the gMS. It is filled by computing the vector $\mathbf{v}^3$ in Fig 3 between the distal end of a segment 2 mapping and some point of the spherical defined by the permissible cone of the 4$^{th}$ segment surface (for the moment, consider it to be the center). However, for an arm with fixed length segments, only a vector of the physical length of segment 3 is allowable. So we can reject any segment 2 hypothesis whose distal locus requires a vector of the wrong length to span the gap. This is a strong pruning condition. It is most strict when the "cone" of segment 4 contains only a single vector. The reader is reminded that the segment 3 vectors are not restricted to the directions available in the quiver. It is a "free vector", $v^3$, determined by the endpoints it must satisfy, but its length is fixed.

$$L^4 \cdot t'^{seg4}_l \left(\mathbf{p}^{end}\right) = L^3 \cdot v^3 \circ L^2 \cdot t^{seg2}_j \circ L^1 \cdot t^{seg1}_i \left(\mathbf{p}^{root}\right) \qquad t^{seg}_j \in T$$

$$L^4 \cdot t'^{seg4}_l \left(\mathbf{p}^{end}\right) = L^3 \cdot v^3 \circ L^2 \cdot t^{seg2}_j \circ L^1 \cdot t^{seg1}_i \left(\mathbf{p}^{root}\right)$$

$$L^3 \cdot v^3 = L^2 \cdot t^{seg2}_j \circ L^1 \cdot t^{seg1}_i \left(\mathbf{p}^{root}\right) - L^4 \cdot t'^{seg4}_l \left(\mathbf{p}^{end}\right)$$

(eqs 2-4)

Note that the inverse notation for segment 4 in eq. 4 denotes the backward direction for that segment.

The pruning conditions before obstacle pruning (described below) for segment 1 are

$$\left| L^1 \cdot t^{seg1}_i \left(\mathbf{p}^{root}\right) - \mathbf{p}^{end} \right| \leq L^2 + L^3 + L^4 \quad \text{for 8DOF}$$

$$\left| L^1 \cdot t^{seg1}_i \left(\mathbf{p}^{root}\right) - \mathbf{p}^{end} \right| \leq L^2 + L^3 \quad \text{for 6DOF}$$

(eqs 5-6)

Once the set of viable segment 1 hypotheses has been restricted, the pruning conditions before obstacle pruning for segment 2 are

$$\left| L^2 \cdot t^{seg2}_j \circ L^1 \cdot t^{seg1}_i \left(\mathbf{p}^{root}\right) - L^4 \cdot t'^{seg4}_l \left(\mathbf{p}^{end}\right) \right| \leq L^3 \quad \text{for 8DOF}$$

$$\left| L^2 \cdot t^{seg2}_j \circ L^1 \cdot t^{seg1}_i \left(\mathbf{p}^{root}\right) - \mathbf{p}^{end} \right| \leq L^3 \quad \text{for 6DOF}$$

(eqs 7-8)

When the target for segment 3 is bounded by a cone of non-zero diameter, the spatial location of the distal end of the segment 2 mapping is not so critical, as illustrated in Fig 2, since that third segment vector need only find some point on the cone-end surface that satisfies its length constraint. Since the quiver subset inside the cone has relatively few members, the cost of checking against all of these, or a sparse sampling is quite low. A faster initial geometric check based on the angle of incidence of the segment 3 vector to the cone central axis and cone diameter can prune even this search. The "free vector" of segment 3 therefore becomes a strong pruning condition on the distal locii of the segment 2 mappings even when the segment 4 cone has some breadth. In practice we allow a small epsilon in the length of the free vector, either because an exact solution is not required, as in path planning, or because we will be able to compute an exact solution at very low cost once we have an good approximation, as will be discussed in the next section.

The "gap" therefore is one of several selection mechanisms for viable IK solutions from the starting sets of segment hypotheses. Simply, a sequence of non-obstructed segment vectors belongs to the IK solution set

$$\left(s_i^{seg1}, s_j^{seg2}, \mathbf{v}^3, s_l^{seg4}\right) \in S_{IK}(p^{root}, p^{end}) \quad \text{where} \quad s^{seg} = L^{seg} \cdot t^{seg}$$

if all its segments survive both the geometric pruning just described, obstacle pruning and joint angle constraints.

**Exact IK solution with discretized segment mappings**

As mentioned above, the epsilon allows a small range of lengths for the segment 3 vector. Once there is a close solution, it can be used to recalculate in a fraction of a microsecond for an exact solution, if required. For a 6DOF solution the endpoint of segment 1 serves as one vertex of a triangle. The target point is the second vertex. The length of the line between them is known and constitutes one side of a triangle whose other sides are the exact lengths of segments 2 and 3. The approximate solution, which has been cleared for collisions, defines the plane of the exact solution, so unless the epsilon was unusually large the exact solution is known to be collision-free as well. The vectors, and subsequently joint angles, can be readily calculated for an exact configuration to the target despite the fact that the first segment's endpoint position was discretized.

For an 8DOF problem the distal end of the second segment serves as the first triangle vertex and the target position as the second. Here, the discretizations of segments 1 and 2 do not preclude an exact solution.

Let $\left(\mathbf{s}_{1,k1}, \mathbf{s}_{2,k2}, \mathbf{v}_3, \mathbf{s}_{4,k4}\right)$ be the inexact solution for 8 DOF. Recalculate for $\mathbf{v}_3$ and $\mathbf{s}_4$ to obtain exact vectors $\mathbf{s}_3$ and $\mathbf{s}'_4$. $\mathbf{s}_{1,k1}, \mathbf{s}_{2,k2}$ remain discretized.

$$\mathbf{p}_2 = \mathbf{p}_0 + \mathbf{s}_{1,k1} + \mathbf{s}_{2,k2}$$
$$\mathbf{d}_3 = \mathbf{p}_t - \mathbf{p}_2$$
$$\mathbf{s}_3 = \mathbf{v}_3 \cdot l3 / |\mathbf{v}_3|$$
$$\mathbf{s}'_4 = \mathbf{d}_3 - \mathbf{s}_3 \quad \text{where } \mathbf{p}_0 \text{ is root and } \mathbf{p}_t \text{ is end}$$

Note that this method for computing an exact solution from a discretized solution works for IK by conventional MSC as well (e.g. for linkages of more than 8DOF).

**Testing for collisions**

Before proceeding to the algorithm, it will be useful to have in mind how obstacles are avoided. As alluded to earlier obstacles are represented as voxels of different value than clear space in a 3D grid. When the mapping for a particular segment hypothesis is being tested for viability in the IK solution set, it is first tested to see if it clears all obstacles. The test is simple. The mapping, or segment hypothesis, is represented as a vector of certain length (i.e. the quiver unit vector times the segment length). That vector is marked along its length by *n* equi-spaced points. The locus of each of those points in the general frame space is used to compute indices into the obstacle grid array. If the voxel at that index set is marked as an obstacle the entire segment hypothesis is rejected as having collided with an obstacle. This is a very low cost approach to checking for collision as it involves no search. In practice, the obstacle volumes are dilated to larger dimensions when the obstacle grid array is initialized to accommodate for the thickness of the arm and some degree of clearance margin (Fig 4) . The intervals of the test points along the segment vector need only be close enough so that the clearance margin guarantees that the arm could not have collided with the actual, as opposed to dilated, boundaries of the obstacle.

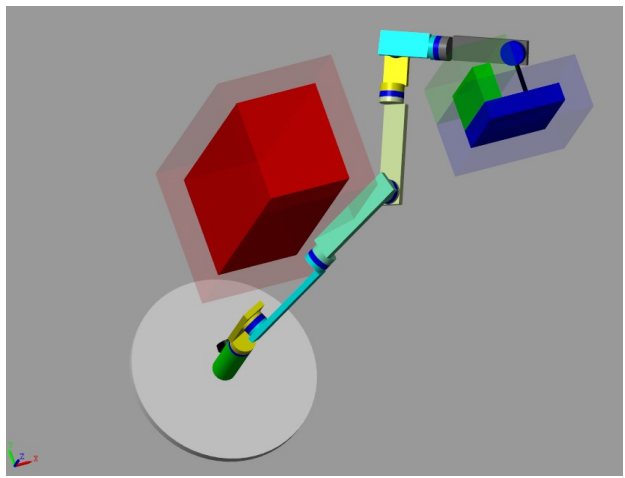

Fig 4: Actual (solid color) and Dilated (translucent) Obstacle Boundaries (frame from SimuLink simulation by Ross Snider). Video at http://www.giclab.com/FastIK_PathPlan.mp4

As will be seen later, the division of a segment into test loci for obstacle collision also provides the waypoints for path planning with no extra computation.

**The procedure for IK solution**

Step 1: All the permissible quiver-generated mappings for segment 1 are tested (a) for joint 1 angle constraints, (b) to determine if the distal end is less than the sum of segment 2 and segment 3 lengths from any of the segment 4 endpoints (or from waypoints during path planning), and (c) for collisions. Only the survivors are marked as viable and saved as a dense set. This step can be done in parallel.

Step 2A: For each viable mapping in the survivor set for segment 1, all the quiver-generated segment 2 mappings from each segment 1 distal locus are checked for (a) joint 2 angle constraints, and (b) obstacle collision.

Step2B: For any survivor from Step 2A the distal endpoint is check for viability across the gap to the $4^{th}$ segment backward mapping endpoints as described above and illustrated in Fig 2. If a viable spanning vector is found, it too is checked for (a) joint 3 angle constraints, and (b) obstacle collisions, in the same manner as the mapped segments, and (c) any self-collision. Any which survive all the pruning conditions constitute a four segment chain which is a viable IK solution.

Step 3: All survivors are captured, and one is chosen as the "reach pose" IK solution. If for some reason, a viable path to this pose cannot be found, as will be discussed below, another reach pose solution from the survivor set may be chosen. This process has yielded the IK solution in vector representation. To control the physical arm, the vector sequence must be converted to joint angles. The elevation angles between segments is computed by the dot product of the segment vectors. The azimuth angles must be computed up the segment chain to obtain the azimuth reference vector for each segment in succession.

Steps 2A and 2B can be parallelized in various ways. For GPGPUs the practical deployment is to create as many threads as there are vectors in the mapping set for segment 2. Each thread computes one segment 2 mapping for all the surviving segment 1 hypotheses from Step 1, and tests those for collision and completion across the gap. Each thread stores any viable solutions it finds.

If this process seems very simple, that is because it is. It is an illustration of the benefit of choosing a well suited representation for solving a problem by computation.

Characterizing this process requires some adjustment in the use of the normal terminology. The algorithm starts to construct a uniformly sampled, discretized subset of *obstacle-constrained configuration space* for the first and second segments. But as it proceeds it prunes this subset on the fly contingent on satisfying the length and joint angle constraints of the segment vector that spans the third segment gap. The surviving segment sequences , $\left(s_i^{seg1}, s_j^{seg2}, \mathbf{v}^3, s_l^{seg4}\right) \in S_{IK}(p^{root}, p^{end})$, constitute a subset of the null space that satisfies (a) obstacle avoidance, and where relevant, (b) constrains the angle of incidence of the end effector axis to the target. This is a restriction of the normal use of *null space* to only configurations in which the end effector is incident to the target location within a specified range of angles.

Pruning on the fly categorizes this as a "greedy" algorithm.

**Path Planning**

The original insight of using MSC as a path planner in the presence of terrain advantages and obstacles was published in [2]. The use of the same path planning process as the guidance for multilink robot was published in [3]. Those examples used linkages of many segments to snake through complex terrain, but in two dimensions. Here the same principle is applied in three dimensions. In the first examples that follow, motion paths are generated as a "by-product" of the computation of the reach pose IK solution. This approach is used as a starting point because it minimizes the computational burden while providing a widely useful method for moving the arm from a standard initial folded pose to the target through challenging obstacles. It is also intended as an illustration of a more general process of path finding, leading to finding motion-efficient paths from arbitrary pose to arbitrary pose, if such paths exist, which will be described afterwards.

The use of the reach pose as the motion path "guide" is not a general solution, but low-cost heuristic that works in many circumstances, specifically where there is sufficient free space next to obstacles near the path to allow the arm to fold in at least one plane which allows the last segment to reach the reach pose defined path, and to swing the folded pose through enough of an angle where the path turns a corner.

These computations could be made for the 8DOF arm, moving the distal end of the $4^{th}$ segment along the path defined by the reach pose. But on the assumption that the $4^{th}$ segment is short, instead the $3^{rd}$ segment distal end traverses the path via a sequence of 6DOF IK solutions and the $4^{th}$ segment is folded out of harm's way by aligning it with the reach pose defined path.

The computation of *n* waypoints along each segment also serves as part of the collision testing. The same *n* points along each segment vector that are tested for collision and survive that test, make ideal waypoints for the path planning stage. All that needs to happen to set this up is for the collision test points to be saved temporarily in an array, and then if the 4 segments survive all tests, for those waypoints to be saved in an array associated with the IK solution. There is virtually no extra cost.

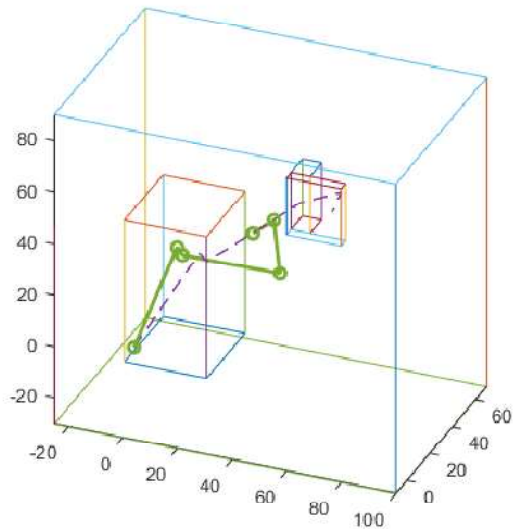

**Fig. 5: Midpath Waypoint IK Solution**

Fig. 5 illustrates an obstacle-clearing IK solution for a midpath waypoint along the lengths of the reach pose shown in Fig 1.

At the last step of the reach pose phase, when an IK solution amongst the surviving set is chosen, the waypoint set associated with it is also selected.

**Smooth sequence computation**

The objective in path planning is that the sequence of arm poses from the initial folded pose to the final reach pose form a sequence yielding smooth continuous arm motion. However, this does not require the precision of endpoint placement required for the reach pose. So a greater epsilon can be tolerated in putting the endpoint (either $3^{rd}$ or $4^{th}$) near the waypoints, while applying some strong criteria so assure that each successive pose along the path is a smooth and nearby transition from the previous.

The most constraining pose is the reach pose. That is, the path pose nearest the reach pose must be a close relative, so the transition from next to last to last does not require big motions. As will be seen, the other end takes care of itself: one of the nice regularities in this solution to the problem.

Since the reach pose is the most constraining, it makes sense to work backward along the waypoints finding the set of IK solutions at each successive waypoint which are closest in configuration to the previously found one. A set of filters based on the previously found pose enforces this consistency. This filtering step will be described below.

A backward sequence of poses is thus captured which, normally, will constitute a smooth path. There are edge cases which don't satisfy the smoothness condition, or may not complete the path at all. But these only occur in highly congested areas which don't allow the arm to fold as needed to follow the path at some point.

The next interesting case arises at the waypoint that coincides with the root locus. Using 6DOF solutions, the distal end of the 3$^{rd}$ segment must be co-located with the root regardless of the backward sequence which took us there. Hence the three segments form a triangle and define a plane (Fig 5). That pose is not the start position of the arm. The arm starting pose is assumed to be folded zig-zag on itself, all segments in a plane (Fig 6).

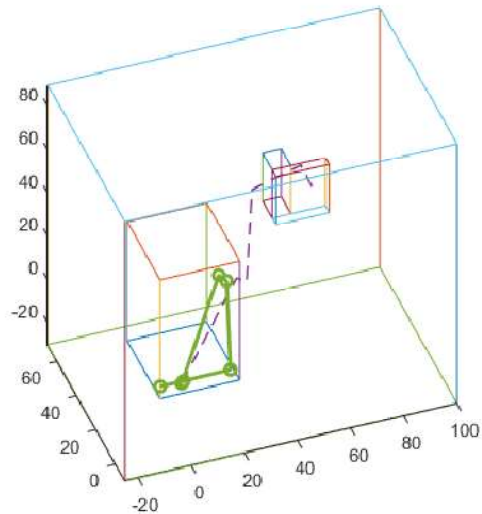

**Fig 5: Root Waypoint IK Solution**

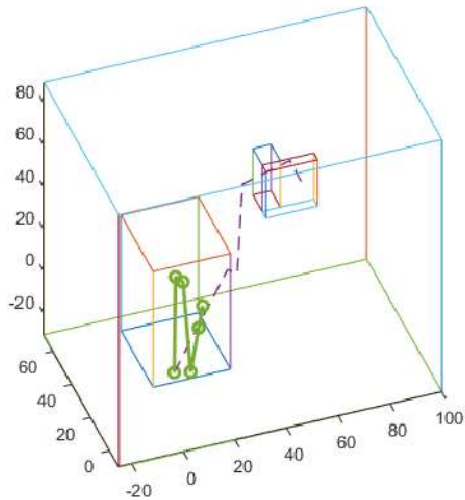

**Fig 6: Initial Folded Arm Pose**

That plane is likely not to be the same plane as the triangle of the root waypoint, but it is a simple matter to rotate the folded pose into the same plane as the root waypoint triangle pose. It requires only rotation of the root joint so that $1^{st}$ segment of the fold pose co-aligns with the $1^{st}$ segment of the root waypoint triangle pose. Now it is a simple matter to unfold the remaining segments from their fold position to the root waypoint pose by angle interpolation. That sequence, reversed completes the backward sequence from reach pose to folded initial pose.

Now executing that whole sequence in forward order constitutes a smooth motion sequence, clearing all obstacles, from initial pose to reach pose

**Filtering for smoothness**

As each IK solution survives both obstacles and completion within a prescribed epsilon of the waypoint it is checked to determine if it is a smooth transition from the previous waypoint solution. There are several tests that work successfully. (a) The distance from the locus of each joint from its corresponding joint in the previous solution must be less than a certain distance approximating to the distance between the waypoints. This is done for joints 1 and 2. When segments 1 and 2 are much longer than segment 3, joint 1 will generally move about half the distance between waypoints along the direction of the waypoints and joint 2 will move about the distance between waypoints. Restricting the allowable motion to approximately these amounts, plus some slack, will prevent dramatic rotations of the solution around the waypoint path. This is a nice geometric regularity that can be exploited with very little cost. Another criterion that can be substituted or added is the change in angle of segment 1 and 2 at each waypoint step, though the distance criterion alone appears to produce smooth motion along the path in all cases tested so far.

There is, however, a caveat regarding the smoothness filtering. When the path makes a large angular turn , the obstruction configuration can require that the pose sequence does in fact require a large rotation as it crosses this turn. If the filtering is set to maximize smoothness in more normal path and obstacle configurations, it can reject a pose solution which effectively turns the corner, but requires a rotation which is precluded.  Fig 8, below is an example, though it is not visible in the frames presented. This circumstance requires that the filter criteria be relaxed for the poses that cross the turn of the path.  This adjustment is best done adaptively (i.e. after failure to find a pose solution at a given waypoint), because the conditions which require it are hard to categorize and detect in advance.  The time penalty for this adaptation is very small: a few milliseconds on large GPU.

Another regularity that may provide speedup in non-parallel implementations is that the geometrically similar solutions tend to cluster nearby in the sequence of computation, so during path planning one need only test mapping indices near those from the previous waypoint to find a smoothly connecting solution for the next waypoint. (This optimization does not appear to provide much benefit in parallel implementations, probably because it increases thread divergence.)

## Other Examples

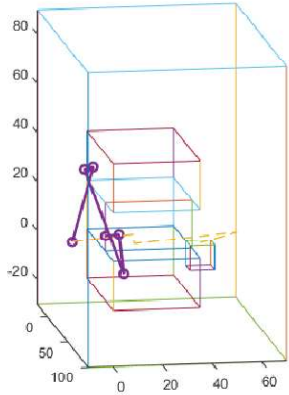 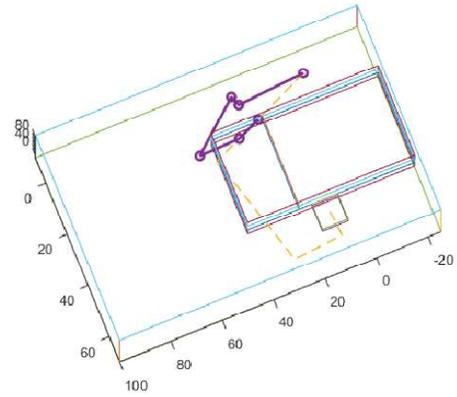

**Fig 7a: Early Path**

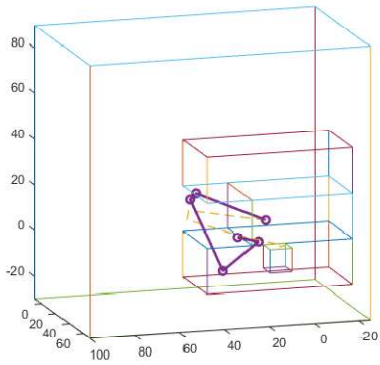 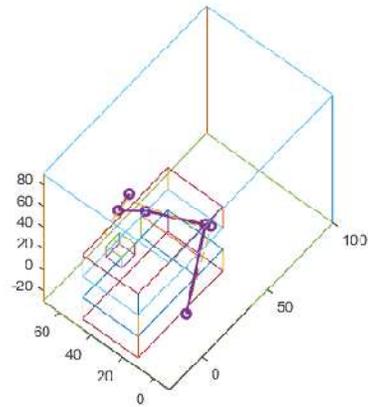

**Fig 7b: Mid Path**

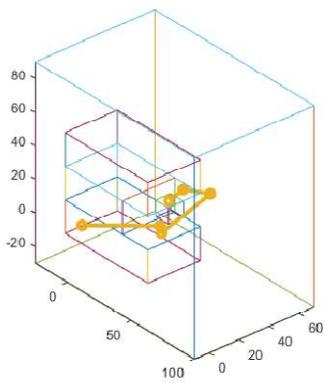 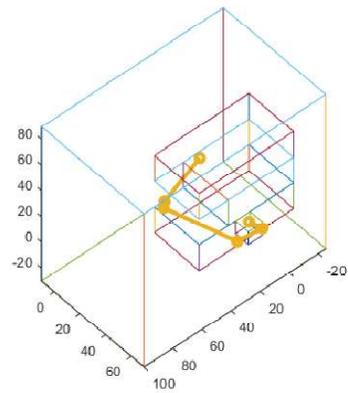

**Fig 7c: Reach Pose, End Path**

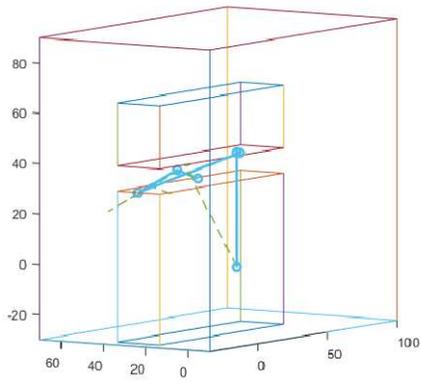 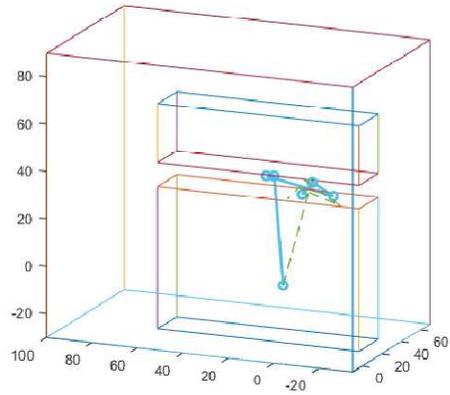

**Fig 8a: Early Path**

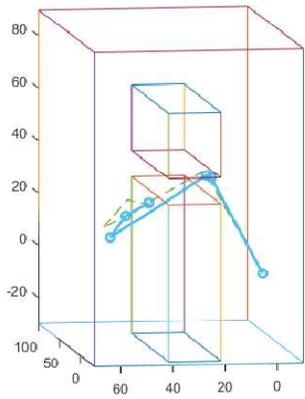 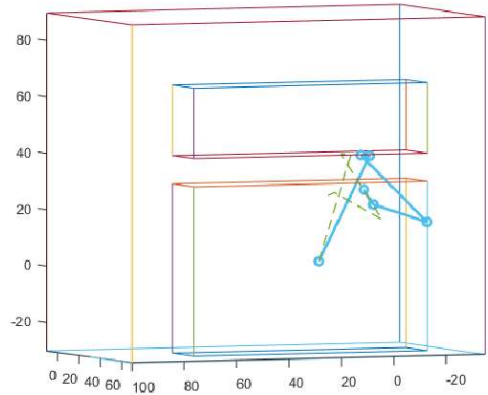

**Fig 8b: Mid Path**

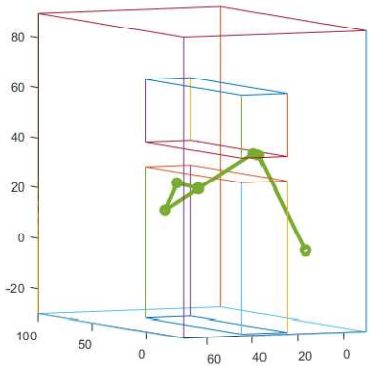 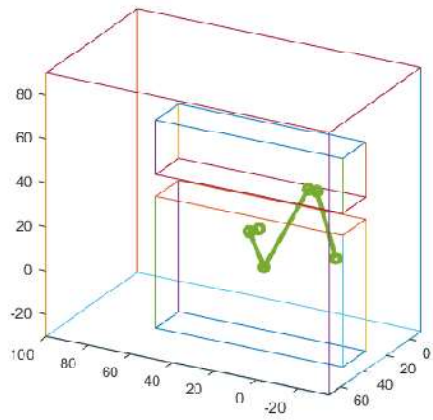

**Fig 8c: Reach Pose, End Path**

**Short Reaches**

It is obvious that for short reaches, e.g. for targets near the root of the arm to targets not requiring the full length of the arm to reach (around obstacles if present), that the use of the reach pose as the motion path leads to unnecessarily long journeys of whichever joint is following the path. A minor modification to the search process for the reach pose IK solution remedies this problem at negligible addition compute cost.

The modification consists of adding another criterion to the search process that produces the reach pose IK solution. This criterion is a near encounter of a segment hypothesis to the target. This is done for all segments (or in practice for the segments 1 and 2 of a 6 or 8 DOF arm). As each segment hypothesis is tested along its length for collision with obstacles, it is also tested for near encounter with the target. If the condition for near encounter to target is satisfied AND no collision has been detected proximal to the near encounter point along the segment, then that segment hypothesis and the point along its length are recorded as a target hit. If this condition is met on a segment 1 hypothesis, that sublength of the segment becomes the shortcut path to the target. It may or may not be part of the reach pose solution ( Figure 9a). If this condition is met on a segment 2 hypothesis, the shortcut path to the target consists of the segment 1 hypothesis from which the satisfying segment 2 hypothesis originates AND the sublength of the segment 2 hypothesis (Figure 9b).

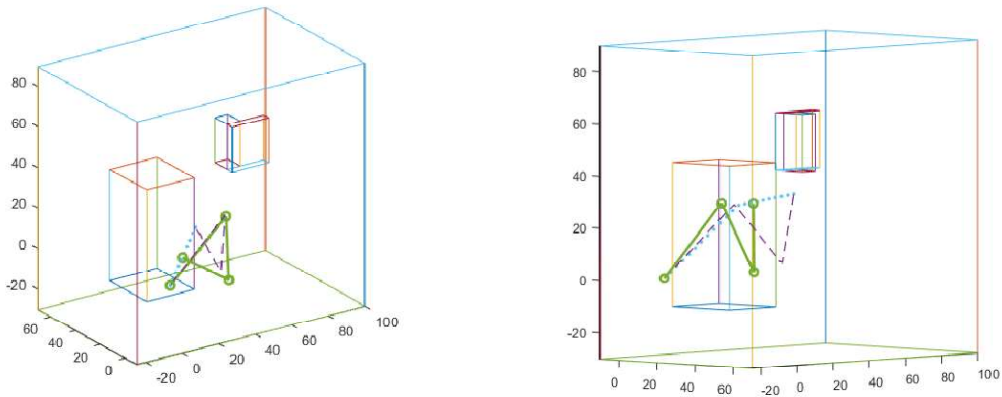

**Fig 9a,b. (a, left) Target reached by Segment 1 sublength . (b, right) Target reached by Segment 2 sublength. Dashed line is reach pose. Dotted line is path.**

If the close encounter condition is not effectively zero then the path maybe completed by either (a) a short bridging vector from the distal end of the satisfying segment hypothesis sublength to the target, or (b) a vector directly from the origin of the satisfying segment hypothesis to the target. In case (b) the new vector will need to be checked for collisions with obstacles. In case (a) if the target is not inside an obstacle it will not need to be checked for collision unless the obstacles have "spikes" which could impinge on the short bridging vector.

By the same procedure as used for the reach pose, waypoints are recorded for those segment hypotheses whose target condition is satisfied. When several have met the condition, the one with the shortest path is selected as the operative path and followed by the same procedure as described for reach pose paths.

If no target condition is met for segments 1 or 2, (or 3 if relevant) then the reach pose is taken as the operative path.

It is not difficult to visualize obstacle shapes which still produce non-optimal paths using the strategy described above. If the bee-line distance to the target is short but obstructed such that a detour much shorter than the arm segment length is required, then the path to the target will still be the length of segment 1 plus some sublength of segment 2. Such cases can be made less frequent by allowing longer bridging vectors.

The more general solution is to use an MSC with as many short or variable length segment layers as necessary to establish a nearly smooth collision-free path, as described in [2,3]. This is also the route to a general solution for paths from one arbitrary pose to another. It involves more computation than the process described above, but should be far less costly than projecting obstacles into a full configuration space and then finding a route through what remains.

**Path between Arbitrary Poses**

The path planning strategy described so far assumes that the arm starts folded such that it can follow the reach path from near the root to the target, or returns along the same path. Two strategies are now discussed for computing paths between arbitrary reachable starting pose and ending or target pose.

The first computes a path for either the end effector or a more proximal joint using the reach path computation method described above with some minor adjustments to the parameters. The easiest way to conceptualize this is to imagine a virtual arm rooted at the end effector locus of the starting pose which then reaches for the target locus, avoiding any obstacles which may be lie between. This virtual arm can have segment lengths appropriate to the likely obstacles to be avoided, and consequently establishes a three or four segment path for the end effector or convenient joint of the actual arm to follow. Of course, there are other constraints on the path selected from the initial solution set. The solution set may contain paths which are not reachable at some points by the actual arm due to length or obstacles. Candidate paths must be tested along its entire length for reachability by the actual arm, constrained by obstacles, until a viable path is found. This can be made efficient by testing only the end waypoints and mid waypoint for each of the three or four segments (or fewer if it is a short reach), and only testing the entire waypoint set once the previous conditions are satisfied.

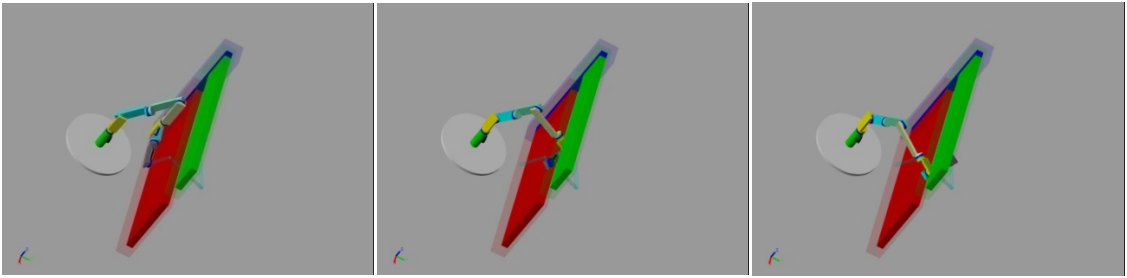

Fig. 10a, b, c: Arbitrary Pose to Arbitrary Pose Path, three waypoints. Frames from video at http://www.giclab.com/ArbitraryStartEnd.mp4

Since the mechanics of computing such paths use the algorithm described earlier no further elaboration of the process is necessary. The path can be computed between more proximal joints (e.g. 6DOF) if the end effector segment can be folded back or otherwise kept out of the way of obstacles during the traverse.

If the approximate length of the maximum path cannot be estimated initially to set the segment lengths, then variable length segments can be implemented to determine both direction and length, just as one would implement a normal reach path IK computation for an arm with prismatic joints in each segment. For extreme obstacle architectures it is possible that no three or four segment path may be viable. In this case a conventional MSC of more than four layers can be used to determine a path of arbitrary complexity, as described for two dimensions in [2] and [3].

Where the complexity of the required direct path imposes too high a computational cost, a less direct but almost always viable out-and-back path can be computed at the cost of just the determining of the reach path and waypoint computation for the ending target pose. One simply reverses the reach path to the starting target, backs the arm into its starting folded pose near the root, and then unfolds on the outward path to the new target. This involves more arm motion, obviously, but is general so long as the path "back" from the starting configuration to the root triangle configuration and the path "out" to the goal configuration are both viable. Depending on obstacle architectures it may be possible to shortcut the route by jumping between back and out paths short of the full folded configuration.

**Dynamic Obstacles**

Consider the situation in which an arm is in motion on a path to a target computed as described above, whether from either a near-root or arbitrary configuration. While the arm is still midcourse a dynamic obstacle is detected in a position where it will constitute a collision-inducing obstacle at a near future configuration of the arm. It is then desirable if a new path can be computed with minimal direction change from the original path that avoids the actual or predicted location of the dynamic obstacle. Assuming that the full traversal of the path would take 3 seconds, each waypoint (for 36 waypoint path) is traversed in approximately 83msec. If the obstacle impinges approximately midcourse, the computation of a new course (on a high end GPU) for somewhat more than half the total path length takes, say (21 + 1)*8.2msec = 180.4msec. Therefore if the collision point is three or more waypoints ahead of the current position of the arm, there is adequate time to compute a new path avoiding the obstacle without stopping the motion of the arm.

The new path must be constrained to lie as close as possible to the current or near future location of the arm along the original path, and pass by the dynamic obstacle with the minimum necessary change of direction.

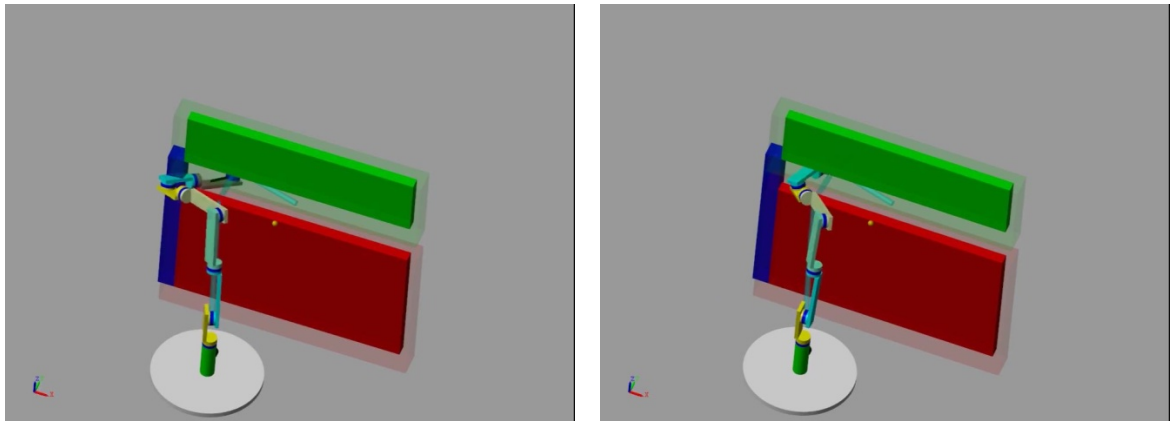

Fig. 11a, b. (Left) arm on original path collides with dynamic obstacle (blue) arriving waypoint 10. (Right) arm on new path starting at waypoint 13 avoids dynamic obstacle (blue) at approximately waypoint 17. Frames from http://www.giclab.com/DynamicCollisionAvoidance.mp4

Assuming the velocity of the dynamic obstacle is such that some diversion of the obstructed path can still reach the target, reach path recomputation by the method described earlier with the dynamic obstacle included with the static obstacles will result in a set of paths to which an objective function can be applied to select the one with the least direction change from the original path. There are a variety of objective functions which can optimize toward that goal, the simplest of which simply minimizes the average distance between the original path and the obstacle avoiding path hypotheses. This selection is still subject to the collision-free viability of all the IK solutions for configurations along the selected path.

The case just described assumes the velocity of the dynamic obstacle is low enough and its shape and course such that the target can still be reached by the arm via an alternate path. Of course there is a class of dynamic obstacle cases for which the target will not be reachable while the dynamic obstacle is still in the way. These cases result in an empty set of solution paths emerging from computation described above.

**Edge and pathological cases**

The method for computing arm IK solutions in the presence of obstacles should always find a reach pose if the architecture of the arm and the configuration of obstacles allows. Nevertheless, having a reach pose and therefore an obstacle free path for some joint of the arm does not mean that there is always a corresponding motion plan. One can readily imagine a configuration of obstacles that would preclude any sequence of articulations from pushing a selected joint along the reach pose defined path. A simple example is a narrow tube around the first or second segments.

Therefore, the method for path planning described so far, while probably widely usable in many circumstances, may not work for some. Either there may be no solution by any means, or there may be viable path plans which cannot be arrived at by the method above.

When no viable IK solution(s) can be found for one or more waypoint along current path hypotheses, the search moves to alternatives:

1) The same path is used, but following it is relaxed. Most likely the early part of the backward solution of waypoint poses will succeed due to similarity to the reach pose. But some waypoint along the sequence will not produce a solution because obstacles will kill all poses that reach to or very near the waypoint. (a) An incremental relaxation of the smoothness filtering and the gap epsilon may allow a solution still reasonably close to the waypoint at the expense of motion smoothness. This procedure may allow a stretch of the reach-pose-guided path to be bypassed until it can be rejoined a number of waypoints down the road. (b) A cloud of target points around the original path may be generated and used as targets, much as described for the segment 4 cone. As solutions are found the cloud would be advanced down the original path.
2) The initial search almost always yields a set of connecting paths. These would normally tested in an order which first tries ones near the original failed one, and if several of these fail, then tests paths far from the ones which have failed.
3) When a path is blocked well along the way to the goal, the arbitrary-to-arbitrary path method can be used to construct a new path from the blocked waypoint to the target. This adds segments to the resulting path which provide more freedom to avoid obstacles. If the detour path is found to be blocked at a waypoint for which no viable IK solution can be found, the arbitrary-to-arbitrary path method can be reapplied from the new blocked waypoint to the goal.

**Joint Angular Velocities**

The division of the path into waypoints provides a simple way to compute the joint angular velocities necessary to produce a desired end effector velocity. The computation described is for controlling motion along the waypoints, so the waypoints and the IK solutions at the waypoints are the only inputs, other than time, to the calculation. To use these calculations to implement the task action after the reach pose, a closely spaced sequence of task waypoints and arm configurations must be provided as inputs. (For tasks that require axial rotation of the final segment or tool, an additional angular term will be needed, as mentioned earlier.)

For a 6 DOF solution, the joint angle representations for the current configuration of the arm, $\mathbf{q}_c$, and $\mathbf{q}_w$ for the next waypoint are defined

$$\mathbf{q}_w = (\theta_1, \phi_1, \theta_2, \phi_2, \theta_3, \phi_3)$$
$$\mathbf{q}_c = (\theta'_1, \phi'_1, \theta'_2, \phi'_2, \theta'_3, \phi'_3)$$

where $\theta$ is azimuth or rotation around axis and $\phi$ is elevation or flexion.

$\mathbf{q}_c$ is queried from the arm joint angle encoders.

The spatial location of the next waypoint, $\mathbf{p}_w$ is known, and the spatial location for joint 4 or the end effector (for a 6 DOF arm), $\mathbf{p}_c$, is computed from $\mathbf{q}_c$ by direct or forward kinematics.

Let the desired average end effector velocity between these waypoint be $v_w$. The time interval to move between the waypoint is

$$t_w = (\mathbf{p}_w - \mathbf{p}_c)/v_w$$

The average joint velocities $\boldsymbol{\omega}_w$ to produce the desired end effector average velocity $v_w$ between the current arm configuration and the waypoint configuration is

$$\boldsymbol{\omega}_w = \left[ (\theta_j - \theta'_j)/t_w, \ (\phi_j - \phi'_j)/t_w \right]_{j=1...n}$$

The actual acceleration will determine what the commanded joint angular velocities must be to produce the computed $\boldsymbol{\omega}_w$ and depending on whether the control objective is time-of-arrival or velocity-on-arrival at waypoint $\mathbf{p}_w$ different adjustments will be needed.

This approach has been implemented on a Kinova Jaco2 7DOF arm. The maximum angular velocity for joints on the arm is 30deg/sec and the position sampling rate is 100Hz. The only effective means of controlling the arm motion is via joint velocities updated as frequently as the basic rate allows. The controlling code recomputes the velocities for all joints multiple times between most waypoints, but the relatively low sampling rate occasionally results in the arm overshooting a waypoint by the time the next position is returned. The adjustment of the new set of joint velocities is always based on the actual arm position and the next waypoint. The resulting motion is continuous and with rare exceptions smooth. Videos of the Kinova arm in motion with the same obstacles as in simulation can be seen at http://www.giclab.com/ObstacleAvoidSimAndJaco2.mp4

Some actuator control APIs which are parameterized by goal joint angles and velocities decelerate the joint angle velocity on close approach to the commanded goal angle. If smooth continuous motion through $\mathbf{p}_w$ is required, new goal angles must be sent to the arm before the deceleration toward $\mathbf{p}_w$ begins. New joint angle velocities are computed from the current configuration $\mathbf{q}_c$ at the moment and the next waypoint $\mathbf{p}_{w+1}$ by the method described above. This strategy is also effective in adaptive motion, where the waypoints or reach pose change in real time. It has been used in emulating insect locomotion over terrain by robot hexapod and works well to produce fluid, natural motion (unpublished).

**Offset Joints and Other Arm Architecture Features**

Figs 3 through 8 depict arms with an offset at joint 2. These were included to demonstrate that the method is not restricted to "coaxial" arm architectures, though the latter is used for description of the method. The offset joint shown rotates in azimuth around the axis of the segment and displaces the elevation rotation at some distance from the segment axis. The implementation of this is straight-forward and involves a single rotation matrix. Further discussion is not appropriate to a presentation of this nature but may be discussed in some detail in a later, more technical presentation. As many joints may be offset as desired, at a small cost in extra computation, approximately 50-80 msec per joint for the whole reach pose and path planning process on a large GPU, although newer GPUs with exposed hardware 4x4 matrix multipliers may reduce this overhead to effectively naught.

**Implementation Performance**

The implementation which generated the examples above is written in C and CUDA (for the GPU executed sections). The first segment computations are done sequentially on the CPU and coded in C. The remaining segment computations are executed in parallel as described earlier on the GPU and are written in CUDA C. 8 waypoints per segment are checked for collision with obstacles, and later used for path planning waypoint computations. On a high-end (200+ watt) Nvidia Quadro GPU the entire computation from start to path takes between 200msec (24 waypoints) and 300msec (36 waypoints) for a coaxial arm and about 250msec (24 waypoints) for a single offset joint arm, as shown in the Figs. The same code takes about a 700msec (36 waypoints) on a low end (45 watt) laptop Nvidia Quadro GPU. The locality-exploiting optimizations described above for serial hardware implementation have not been tested, since the code diverged from all-serial implementation early in development.

Further MSC references and tutorials can be found at www.giclab.com

Videos for the examples in the text can also be found in the References/Technical Reports section of www.giclab.com.